# Classification of Radiologically Isolated Syndrome and Clinically Isolated Syndrome with Machine-Learning Techniques

Virginia Mato-Abad,[1] Andrés Labiano-Fontcuberta,[2] Santiago Rodríguez-Yáñez,[1] Rafael García-Vázquez,[1] Cristian R Munteanu,[3, 4] Javier Andrade-Garda,[1] Angela Domingo-Santos,[2] Victoria Galán Sánchez-Seco,[2] Yolanda Aladro,[5] Mª Luisa Martínez-Ginés,[6] Lucía Ayuso,[7] Julián Benito-León,[2, 8, 9] *

From the ISLA,[1] Computer Science Faculty, A Coruna University, A Coruña, Spain; Department of Neurology,[2] University Hospital "12 de Octubre", Madrid, Spain; RNASA-IMEDIR,[3] Computer Science Faculty, A Coruna University, A Coruña, Spain; Biomedical Research Institute of A Coruña (INIBIC)[4], University Hospital Complex of A Coruña (CHUAC), A Coruña, Spain; Department of Neurology,[5] Getafe University Hospital, Getafe, Spain; Department of Neurology,[6] University Hospital "Gregorio Marañón", Madrid, Spain; Department of Neurology,[7] University Hospital "Principe de Asturias", Alcalá de Henares, Spain; Centro de Investigación Biomédica en Red sobre Enfermedades Neurodegenerativas (CIBERNED),[8] Spain; Department of Medicine,[9] Complutense University, Madrid, Spain.

**Abstract.**

**Introduction:** The unanticipated magnetic resonance imaging (MRI) detection in the brain of asymptomatic subjects of white matter lesions suggestive of multiple sclerosis (MS) has been named as radiologically isolated syndrome (RIS). As the difference between early MS (i.e., clinically isolated syndrome [CIS]) and RIS is the occurrence of a clinical event, it should be logical to improve detection of subclinical form without interfering with MRI as there are radiological diagnostic criteria for that. Our objective was to use machine-learning classification methods to identify morphometric measures that help to discern patients with RIS from those with CIS.

**Methods:** We used a multimodal 3T MRI approach by combining MRI biomarkers (cortical thickness, cortical and subcortical grey matter volume, and white matter integrity) of a cohort of 17 RIS and 17 CIS patients for single-subject level classification.

**Results:** The best proposed models to predict the CIS and RIS diagnosis were based on the Naive Bayes, Bagging and Multilayer Perceptron classifiers using only three features: the left rostral middle frontal gyrus volume, and the fractional anisotropy values in the right amygdala and in the right lingual gyrus. The Naive Bayes obtained the highest accuracy (overall classification, 0.765 and AUROC, 0.782).

**Conclusions:** A machine-learning approach applied to multimodal MRI data may differentiate between the earliest clinical expressions of MS (CIS and RIS), with an accuracy of 78%.

*\*Corresponding author: Julián Benito-León. Av. de la Constitución 73, portal 3, 7º Izquierda, 28821 Coslada, Madrid, Spain. E-mail: jbenitol67@gmail.com

**Competing interests**

The authors declare no competing financial interests.

**Authors Roles:**

Dr. Virginia-Mato (virginia.mato@udc.es) collaborated in: 1) the conception, organization of the research project; 2) the statistical analysis design, and 3) the writing of the manuscript first draft and the review and critique of the manuscript.

Dr. Labiano-Fontcuberta (gandilabiano@hotmail.com) collaborated in: 1) the conception, organization and execution of the research project; 2) and the review and critique of the manuscript.

Dr. Rodríguez-Yáñez (santiago.rodriguez@udc.es) collaborated in: 1) the conception,

organization of the research project; and 2) the review and critique of the manuscript.

Dr. García-Vázquez (rafael.garcia@udc.es) collaborated in: 1) the conception, organization of the research project; and 2) the review and critique of the manuscript.

Dr. Munteanu (c.munteanu@udc.es) collaborated in: 1) the conception, organization of the research project; and 2) the review and critique of the manuscript.

Dr. Andrade-Garda (javier.andrade@udc.es) collaborated in: 1) the conception, organization of the research project; and 2) the review and critique of the manuscript.

Dr. Domingo-Santos (gela_yo@hotmail.com) collaborated in: 1) the conception, organization of the research project; and 2) the review and critique of the manuscript.

Dr. Galán Sánchez-Seco (vickyg_s@hotmail.com) collaborated in: 1) the conception, organization of the research project; 2) the statistical analysis design; and 3) the review and critique of the manuscript.

Dr. Aladro (yolanda.aladro@salud.madrid.org) collaborated in: 1) the conception, organization of the research project; and 2) the review and critique of the manuscript.

Dr. Martínez-Gines (marisamgines@hotmail.com) collaborated in: 1) the conception, organization of the research project; and 2) the review and critique of the manuscript.

Dr. Ayuso (layusoperalta@gmail.com) collaborated in: 1) the conception, organization of the research project; and 2) the review and critique of the manuscript.

Dr. Benito-León (jbenitol67@gmail.com) collaborated in: 1) the conception, organization of the research project; and 2) the review and critique of the manuscript.

**Disclosures:**

Dr. Mato-Abad reports no disclosures.

Dr. Labiano-Fontcuberta reports no disclosures.

Dr. Rodríguez-Yáñez reports no disclosures.

Dr. García-Vázquez reports no disclosures.

Dr. Monteanu reports no disclosures.

Dr. Andrade-Garda reports no disclosures.

Dr. Domingo-Santos no disclosures.

Dr. Galán Sánchez-Seco reports no disclosures.

Dr. Aladro reports no disclosures.

Dr. Martínez-Ginés reports no disclosures.

Dr. Ayuso reports no disclosures.

Dr. Benito-León reports no disclosures.

**Acknowledgments and Funding**

This research was partially supported by "Collaborative Project in Genomic Data Integration (CICLOGEN) (PI17/01826), granted by the Spanish Health Research Agency from the National Plan for Scientific and Technical Research and Innovation 2013–2016 and FEDER Funds. Dr. Benito-León is supported by the National Institutes of Health, Bethesda, MD, USA (NINDS #R01 NS39422), the Commission of the European Union (grant ICT-2011-287739, NeuroTREMOR), the Ministry of Economy and Competitiveness (grant RTC-2015-3967-1, NetMD—platform for the tracking of movement disorder), and the Spanish Health Research Agency (grant PI12/01602 and grant PI16/00451).

**INTRODUCTION**

The steady increase in the use of magnetic resonance imaging (MRI) for the evaluation of different medical conditions, such as headaches or dizziness, has led to the emergence of a new entity named radiologically isolated syndrome (RIS), which is characterized by incidental brain MRI finding of white matter lesions demonstrating dissemination in space in subjects with a normal neurologic examination, and without historical accounts of typical multiple sclerosis (MS) symptoms.[1]

Multiple sclerosis (MS) is an inflammatory, demyelinating, and neurodegenerative disease that predominantly affects young adults and it is characterized by heterogeneous manifestations and evolution. At the clinical onset of the disease, approximately 85% of patients experience an acute or subacute episode of neurologic disturbance, known as clinically isolated syndrome (CIS).[2]

The differentiation of RIS from CIS, typically the earliest clinical expression of MS, may be challenging. In fact, a number of recent studies suggest that RIS and relapsing-remitting MS or CIS patients share both nonmotor clinical features[3, 4] and quantitative brain tissue damage,[5] thereby suggesting that RIS, as an entity, may reflect the earliest and preclinical form of MS. Although the difference between CIS and RIS, by definition, is the presence of clinical symptoms, some people with CIS may have symptoms so mild that they go unnoticed and hence they could be diagnosed with RIS. From a clinical point of view, it is extremely important to differentiate both entities (RIS vs. CIS), since CIS patients are usually treated with disease-modifying therapies, unlike RIS patients.[6]

In this context, several quantitative MRI methods have been established to assess changes in brain areas, which appear to be normal on conventional MRI, the so-called normal-appearing grey and white matter.[7] Also, the width of the cortical grey matter layer that covers the surface of the brain, referred to as cortical thickness, has been assessed as an useful measure in a variety of disorders to study neuroanatomical patterns, including MS.[8] Diffusion tensor imaging (DTI) has the potential to quantify microstructural changes that modify the integrity of brain tissues.[9] By using DTI, the brain tissue microstructure can be determined by quantitative indexes, such as mean diffusivity (MD), which is affected by cellular size and integrity, and fractional anisotropy (FA), which reflects the degree of alignment of cellular structures within fibre tracts and their structural integrity.[9]

The above MRI modalities provide extremely high-dimensional raw data like cortical thickness, cortical and subcortical grey matter volumes or FA and MD values. The analysis of these biomarkers, using statistical packages for neuroimaging analysis like SPM (www.fil.ion.ucl.ac.uk/spm/), FSL (fsl.fmrib.ox.ac.uk/fsl/fslwiki) or FreeSurfer (surfer.nmr.mgh.harvard.edu), allows us to study differences between groups (e.g. RIS vs. CIS). However, these methods are not applicable on a single-subject level and therefore do not improve the clinical diagnosis potential. To overcome this issue, machine-learning techniques have recently been identified as promising tools in neuroimaging data analysis for individual class prediction.[10] Automatic classification techniques provide tools for analysing all these variables simultaneously and observe inherent disease-related patterns in the data.[10]

Despite the fact that machine-learning techniques have been widely used for MRI images in several neurological and neurodegenerative disorders, including MS for predicting disease course,[11] classifying between different MS disease courses,[12] or even for predicting CIS conversion to MS,[13] no study to date has been conducted to discern between CIS and RIS patients. We hypothesized that a machine-learning approach, applied to multimodal MRI data, can differentiate between CIS and RIS patients. The aim of this study was therefore to test and evaluate the effectiveness of machine-learning schemes for single-subject level classification of individuals affected by earliest forms of MS (CIS and RIS). Towards this purpose, we used a multimodal 3T MRI approach by combining MRI biomarkers (cortical thickness, cortical and subcortical grey matter volume, and white matter integrity) of a cohort of RIS and CIS patients. The collection of WEKA machine-learning algorithms was used for this purpose.

## METHODS

### *Participants*

Seventeen RIS patients (13 women, 4 men; mean age 41.6 years, range 27–52 years) were recruited at four centres specialized in demyelinating diseases in Madrid (Spain). These patients had detected after undergoing conventional brain 1.5 T MRI for various medical events not suggestive of MS. RIS patients who were included in the current study represent a subset of our previous RIS cohort;[3, 4] specifically, only those who underwent a complete 3 T multimodal MRI study.

Brain white matter abnormalities were initially identified by a neuroradiologist and subsequently examined by an MS specialist at each clinical site to guarantee the diagnostic criteria for RIS by Okuda et al.[14]

Seventeen patients (12 women, 5 men; mean age 39.5 years, range 30–55 years) who had presented with a CIS suggestive of MS were recruited from the University Hospital "Gregorio Marañón," and from the University Hospital of Getafe, both in Madrid (Spain). All CIS patients underwent a complete neurological evaluation, including Expanded Disablity Status Scale (EDSS)[15] by experienced neurologists (M.L.M.-G. and Y.A). Patients selected met the following inclusion criteria: 1) single clinical episode indicative of MS; 2) total follow-up time of at least 3 months from the occurrence of the first inflammatory demyelinating event; and 3) the presence of ≥ 1 asymptomatic T2 lesion(s) in at least two or more brain locations considered characteristic for MS (juxtacortical, periventricular, infratentorial, and spinal cord)[16] at the initial or follow-up MRI. Participants were excluded if they had received steroid medication during the month before the study inclusion and a longitudinal evaluation longer than five years.

All patients (CIS and RIS) underwent a multisequence MRI examination, which was acquired in a single session using a single 3 T scanner at CIEN (*Centre for Research on Neurological diseases, in Spanish*) Foundation in Madrid (Spain).

All the participants included in the study gave their written informed consent after full explanation of the procedure. The study, which was conducted in accordance with the principles of the Helsinki declaration of 1975, was approved by the ethical standards committee on human experimentation at the University Hospital "12 de Octubre" (Madrid).

*Measurement Instruments*

*MRI Acquisition*

All MRI data were acquired with a clinical 3T Signa HDx MRI scanner (GE Healthcare, Waukesha, WI) using an 8-channel phased array coil. The imaging (MRI) standardized protocol (without injection of contrast agent) included a 3D T1-weighted SPGR with a TR=10.012 ms, TE=4.552 ms, TI=600 ms, NEX=1, acquisition matrix=288×288, full brain coverage, resolution=0.4688×0.4688×1 mm, flip angle=12. The diffusion-weighted image (DWI) protocol acquisition consisted of 3 images without diffusion gradients (b=0 s/mm2) followed by 45 images measured with 45 directions (b=1000 s/mm2) isotropically distributed in space. Additional parameters of the acquisition were: TE=85.3 ms, TR=10.100 ms, flip-angle=90, slice thickness=3 mm (no gap), resolution=2.6042×2.6042×2.6 mm, FOV=250 mm and axial acquisition.

All MRI acquisitions and image postprocessing (see below) were performed by a neuroradiologist (JA-L, see acknowledgments) and a physicist (VM-A) who were blinded to the clinical diagnoses.

*Data Post-processing*

*MRI*

Cortical thickness and cortical volume measurements were calculated using the freely available software FreeSurfer (http://surfer.nmr.mgh.harvard.edu/). Using a surface-based approach, FreeSurfer can automatically segment the brain into different cortical regions of interest and calculate average thickness in the defined regions. In brief, images underwent pre-processing including intensity normalization and skull stripping, which was followed by labelling of cortical and subcortical regions. FreeSurfer's main cortical reconstruction pipeline began with the registration of the structural volume with the Talairach atlas.[17] After bias field estimations and the removal of these bias, the skull was stripped and subcortical white and grey matter structures were segmented.[18] Next, tessellation, automated topology correction, and surface deformation routines to create the white/grey (white) and grey/cerebrospinal fluid (pial) surface models.[19] These surface models were then inflated, registered to a spherical atlas, and used to parcellate the cortical mantle, according to gyral and sulci curvature.[20] The closest distance from the white surface to the pial surface at each surface's vertex was defined as the thickness.[20] Average cortical thickness, surface area, and total volume statistics corresponding to each parcellated region were then computed. The accuracy of FreeSurfer's results was then assessed visually for the different participants.

*DWI*

DWI data were pre-processed with FMRIB's Diffusion Toolbox (fsl.fmrib.ox.ac.uk/fsl/fslwiki/FslOverview). Pre-processing consisted of eddy-current correction, motion correction, and the removal of non-brain tissue using the robust Brian Extraction Tool.[21] Diffusion tensor images (DTIs) were created using the weighed least squares fitting method. We derived images of FA or MD from the DTI. To calculate the specific FA or MD values in the cortical and subcortical regions, we needed to map the DTI space to the Freesurfer's structural space (described above). The FA or MD maps were resampled by means of a rigid-body transformation from the diffusion to the structural space. After that, the FA or MD mean and standard deviation values from the Freesurfer's subcortical regions were computed.

*Dataset and analyses description*

The dataset was divided into the previously described groups (RIS and CIS). A total of 306 attributes resulted from the post-processing analysis for each of the 34 cases/examples: the volume, the FA and MD values corresponding to 66 cortical[18] and 14 subcortical[20] parcellated regions obtained with FreeSurfer as well as the average cortical thickness of the 66 cortical regions. Imaging features were coded using the following nomenclature: *cerebralHemisphere_cerebralRegion_feature* where "*cerebralHemisphere*" can be *LH* (for left hemisphere) or *RH* (for right hemisphere); "*cerebralRegion*" takes the FreeSurfer atlas coding and "*feature*" can be *volume* (for volumetric measures), *thickness* (for cortical thickness measures), *FA* (for fractional anisotropy measures) or *MD* (for mean diffusivity measures).

The WEKA data mining software was used to search the best classification models using different sets of data and machine-learning techniques.[22] The calculations tried models using attributes from both hemispheres. All the WEKA algorithms were applied using the well-known 10-fold cross-validation technique to split the data. The performance of prediction models for a two-class problem was evaluated using a confusion matrix. There are several numbers of well-known accuracy measures for a two-class classifier in the literature. In the current study, we present the true positives (TP) score and the AUROC values for the validation process. Previous studies have suggested that, in general, AUROC is the best measure for model comparison.[23]

Feature selection (FS) is a helpful stage prior to classification for dimensionality reduction, as well as selecting proper features and omitting improper features.[24] The aim of feature selection is to select a subset of extracted variables to reduce the number of input variables for the classifier, since the number and relevance of the input variables can affect the performance of the model.[24] A FS approach was performed to find the subset of features best describing the structure of the data. The FS process can answer the main question of how many and which are the most important features to discriminate between CIS and RIS patients. There are mainly three approaches for FS: Filter, Wrapper and Embedded. For the current study, a Filter approach was used to assess the relevance of features by looking only at the intrinsic properties of the data as a fast and simple computationally approach.

The WEKA data mining software contains a collection of algorithms, divided into different family groups, to solve classification problems. They have all been tested in the current study, but the ones that have been proven relevant are the following:

- Bayes Theorem, such as the Naive Bayes Classifier, which depending on the hypothesis, the presence or absence of a disease is independent of the feature space. This classifier has been used to solve different problems in the medical field, such as data analysis, prediction models or improving the brain diagnosis accuracy by means of MRI images.[25]
- Estimation of functions, such as the Artificial Neural Networks (ANNs), which are flexible, non-linear and multidimensional mathematical systems capable of solving complex functions in very diverse fields. They have been widely used in the medical area, such as Alzheimer's disease and mild cognitive impairment diagnosis, by combining different MRI techniques,[26] or for predicting the short-term prognosis of MS,[25], among others.

- Combination of multiple algorithms. An example of this kind of algorithms is Bagging (stands for Bootstrap Aggregation), which combines the results of base classifiers treating each model with equal weight to generate final prediction. To generate better prediction models, each base classifier was trained using randomly drawn sample sets (bootstrap samples) with replacement from original training set.[27]

## RESULTS

### Sample characteristics

Table 1 summarizes the demographic and clinical characteristics of the entire sample and shows that the groups were well-matched for age ($F(2.48)=0.68$, $p=0.51$), sex ratio ($\chi2=0.55$, $p=0.75$), and educational level ($\chi2=0.08$, $p=0.95$). Reasons for the first RIS patients MRI, which was performed a mean of 4.1 years (range 1-11) earlier, were: headache (N=5); dizziness (N=4); tinnitus-hypocusia (N=3); syncope (N=1); restless legs (N=1); research control (N=1); traffic accident (N=1); and prolactinoma (N=1).

CIS patients were characterized by low clinical disability (median EDSS score of 0, range 0-4) and a relatively short duration of disease (median disease duration from clinical onset = 12 months). Only 3/17 (17.6%) CIS patients had a disease evolution from clinical onset shorter than 12 months, whereas 7/17 (41.1%) had disease duration from symptom onset of at least 48 months. Of the 17 CIS patients, five (29.4%) presented with spinal cord symptoms, four (23.5%) with optic neuritis; three (17.6%) with brainstem symptoms; three with polysymptomatic onset and two (11.8%) with hemispheric cerebral symptoms. All CIS subjects fulfilled dissemination in space according to the McDonald 2010 criteria.[28]. Four (23.5%) were on disease-modifying treatment (interferon-beta).

Using a cut-off score of 8 on the Hamilton Depression Rating Scale total score,[29] nine (52.9%) of the RIS group had at least mild clinical depression, with more than half of them (N=5, 55.9%) having moderate depressive symptoms, rates identical to that observed among CIS patients.

Table 2 shows the results of RIS and CIS classification for those WEKA's machine-learning algorithms depicting statistic value greater than 0.7. The models were based on MRI biomarkers for both hemispheres (LH+RH) and on individual hemispheres (LH or RH) for all possible classification problems. MRI biomarkers were identified by the nomenclature described above.

The best proposed models to predict the CIS and RIS diagnosis were based on the Naive Bayes, Bagging and Multilayer Perceptron (MLP, one of the most popular ANN) classifiers using only three features: the left rostral middle frontal gyrus volume, and the FA values in the amygdala and in the lingual gyrus, both in the right hemisphere. The Naive Bayes obtained the highest accuracy (overall classification, 0.765 and AUROC, 0.782). These methods could classify above 70% of accuracy using only the information provided by two features (FA values in the amygdala and in the lingual gyrus). Only the FA value in the right amygdala achieved classification results above the 70% with the Naive Bayes Classifier and the MLP.

**DISCUSSION**

Clinicians usually use conventional MRI technology more often than novel techniques. Efforts to improve the characterization of RIS are not only essential to prevent overdiagnosis, but to enhance medical counselling, surveillance recommendations, and future treatment strategies.

The current work presents for the first time the classification of RIS and CIS with machine-learning techniques using multimodal 3T MRI data. Our analysis of MRI and DWI biomarkers showed that only three features were relevant for the classification process: the volume of the left rostral middle frontal gyrus and the FA values in both the right amygdala and the right lingual gyrus (overall classification, 0.765 and AUROC, 0.782 for the Naive Bayes). We have not found any previous study that relates changes in lingual gyrus with RIS. However, changes in frontal gyrus and in the amygdala are in line with clinical findings found in previous studies. For example, RIS patients have higher rates of depression, particularly anxious depression when compared with CIS patients.[3] This kind of depression is associated with modulation of amygdala connectivity.[30] Specifically, the right amygdala has been shown to be the most significant feature to discriminate CIS and RIS groups in our study. In addition, it has been suggested that patients with major depression may have right hemispheric dominant pre-attentive dysfunction.[31] Accordingly, if RIS patients tend to present higher rates of depression, they should have more affected the right hemisphere, which agrees with our results. Regarding the middle frontal gyrus, a recent study has found assessed that RIS patients show altered microstructural integrity in bilateral frontal sub-gyral regions.[5]

Among all the methods of the WEKA data mining software, the Naive Bayes classifier is the one that achieved the better performance. This classifier is widely recognized as a simple and effective probabilistic classification method for biomedical applications.[32] Bayesian methods have been able to account for a range of inference problems relevant to biomedical applications. The analysis and classification of biomedical data is challenging because it must be done in the face of uncertainty (datasets are often noisy and with missing data). Bayesian decision theory is the principal approach for inferring underlying properties of data in the face of such uncertainty.

Our results also confirm that Bagging and MLP are accurate techniques for predicting CIS and RIS diagnosis. Bagging algorithm is mainly useful in the case of small sample sizes and high dimensional datasets to get a more robust estimate.[33] On the other hand, there are several studies across different fields that show that ANN approaches, MLP in particular, provide high prediction accuracies to solve classification problems [26]

Although different studies in predicting disease course in MS have been undertaken,[34, 35] there remains much to learn about machine-learning techniques in demyelinating diseases of the central nervous system, especially in RIS. Therefore, our results cannot be compared with previous studies. Further, the comparison with other studies in MS is difficult, since they report a wide range of different accuracies for classification and prediction tasks and they have used different cohorts, features, and techniques. The classification accuracy could be influenced by several factors including both methods and cohort properties. Feature extraction methods, feature selection or classification tools, image quality, number of subjects, demographics and clinical diagnosis criteria are also important considerations. Despite all this, the scores obtained in the current study agree with previous ones based on machine-learning techniques in MS, where the accuracy tends to be around 70%. However, the studies published to date in this field are mainly focused in predicting disease course in MS patients and, in the current study, we are dealing with individuals affected by the earliest forms of MS, RIS and CIS, trying to discriminate between these two different clinical conditions. For example, Wottschel et al.,[13] predicted clinical conversion to MS from 74 CIS patients during one- and three-year follow-up using support vector machines with an accuracy of 71.4% and 68% for one- and three-years, respectively using MRI data and clinical information. Also, Bejarano et al.,[25] explored different WEKA algorithms (Naive Bayes, simple logistic, decision trees, and MLP) for predicting short-term disease outcomes in MS in a cohort of 51 CIS patients using clinical, imaging, and neurophysiological variables. In this study,[25] they found that the MLP yielded the better performance for predicting the EDSS change two years later with and accuracy around 80%. The study carried out by Zhao et al.,[11] explored different machine learning methods for predicting MS disease course of a total of 1,693 subjects (574 with MRI data) belonging to *The Comprehensive Longitudinal Investigation of MS at the Brigham and Women's Hospital* (CLIMB). Their main outcome in these experiments was to assess the ability of clinical and MRI features to predict EDSS status at up to five years.[34] The purpose of Ion-Mărgineanu et al.,[12] was to classify 87 MS patients in the four clinical forms defined by the McDonald criteria, using machine-learning algorithms on clinical data, lesion loads

and MRI metabolic features. They found scores between 70 and 85% depending on the different groups.[12]

The study should be interpreted within the context of several limitations. The most important, the small sample size. Given the low prevalence and incidence of the disease, the RIS literature generally comprises studies with small sample sizes.[1] However, we could classify with an overall accuracy of 78% between CIS and RIS patients even with these small numbers. Notwithstanding, it would be important to replicate these findings in a larger and independent data set. Also, white matter lesions were not in-painted before submitting the image to FreeSurfer, which could compromise the accuracy of measured cortical thickness. However, lesion in-painting has only a small effect on the estimated regional and global cortical thickness.[36] Further, we have previously reported that white matter lesion volume did not differ significantly between RIS and CIS groups [5].

In closing, we have shown that a machine-learning approach applied to multimodal MRI data may differentiate between the earliest clinical expressions of MS (CIS and RIS), with accuracy of 78% (AUROC, 0.782). We used a multimodal approach by combining MRI biomarkers (cortical thickness, cortical and subcortical grey matter volume, and white matter integrity) over the collection of WEKA machine-learning algorithms. Our study reflects that the best results in terms of AUROC are the Naive Bayes, Bagging and MLP classifiers. Although this is the first application of machine-learning techniques to the classification of RIS patients, the scores obtained are in accordance with previous MS studies. This technique has the potential to be used to research in demyelinating diseases.

**Table 1**: Demographic and clinical characteristics of the patients.

|  | Clinically isolated syndrome patients (N=17) | Radiologically isolated syndrome patients (N=17) |
|---|---|---|
| **Female / male (ratio)** | 12 / 5 (2.4) | 13 / 4 (3.2) |
| **Age in years** | 39.5 ± 6.1 (30-55) | 41.6 ± 7.1 (27-52) |
| **Educational level** | | |
| *Primary studies* | 5 (29.4%) | 4 (23.5%) |
| *Secundary studies* | 4 (23.5%) | 7 (41.2%) |
| *University studies* | 8 (47.1%) | 6 (35.3%) |
| **Mean age at RIS diagnosis** | - | 37.7 ± 7.4 |
| **≥ 9 T2 white matter lesions (%)** | 10 (58.8%) | 6 (35.3%) |
| **Spinal cord lesions on MRI** | 10 (58.8%) | 17 (100%) |
| **Gadolinum-enhancing lesions** | 0 (0%) | 3 (17.6%) |
| **Dissemination in time criteria** | 4 (23.5%) | 5 (29.4%) |
| **Dissemination in space Barkhof criteria (%)** | 11 (64.7%) | 15 (88.2%) |
| **Dissemination in space Swanson criteria (%)** | 17 (100%) | 17 (100%) |
| **Expanded Disability Status Scale total score**[+] | 0 (0-4) | - |

Mean ± standard deviation (range) and frequency (%) are reported.
[+]Ordinal variables or variables that were not normally distributed and therefore median (interquartile range) is provided.
[a] % of patients with clinically significant depression;

**Table 2:** Results for CIS-RIS classification using different sets of LH and RH brain areas for the WEKA algorithms with AUROC value greater than 0.7

| LH + RH: lh_rostralmiddlefrontal_volume, rh_Amygdala_FA, rh_lingual_FA | | | | |
|---|---|---|---|---|
| *ML method* | *TP score for CIS* | *TP score for RIS* | *Overall accuracy* | *AUROC value* |
| Naïve Bayes | 0.824 | 0.706 | 0.765 | 0.782 |
| Multilayer Perceptron | 0.824 | 0.647 | 0.735 | 0.720 |
| Bagging | 0.647 | 0.765 | 0.706 | 0.761 |
| **RH: rh_Amygdala_FA, rh_lingual_FA** | | | | |
| *ML method* | *TP score for CIS* | *TP score for RIS* | *Overall accuracy* | *AUROC value* |
| Naïve Bayes | 0.706 | 0.647 | 0.676 | 0.727 |
| Bagging | 0.765 | 0.706 | 0.735 | 0.754 |
| **RH: rh_Amygdala_FA** | | | | |
| *ML method* | *TP score for CIS* | *TP score for RIS* | *Overall accuracy* | *AUROC value* |
| Naïve Bayes | 0.706 | 0.706 | 0.706 | 0.711 |
| Multilayer Perceptron | 0.824 | 0.706 | 0.765 | 0.730 |

LH, left hemisphere; RH, right hemisphere; FA, fractional anisotropy; ML, machine-learning; TP, True Positives.